\documentclass[12pt]{article}
\usepackage{multirow}
\usepackage[unicode=true]{hyperref}
\usepackage{breakurl}
\usepackage{pifont}
\usepackage{amsmath,amsfonts,stmaryrd,amssymb} 
\usepackage{ifthen}
\usepackage{verbatim}
\usepackage{graphicx}
\usepackage{color}
\PassOptionsToPackage{normalem}{ulem}
\usepackage{ulem}
\usepackage{url}

\usepackage{enumerate} 

\usepackage[ruled]{algorithm2e} 

\usepackage[framemethod=tikz]{mdframed} 

\usepackage{listings} 
\usepackage{geometry} 

\usepackage[utf8]{inputenc} 
\usepackage[T1]{fontenc} 

\usepackage{XCharter} 

\mdfdefinestyle{commandline}{
	leftmargin=10pt,
	rightmargin=10pt,
	innerleftmargin=15pt,
	middlelinecolor=black!50!white,
	middlelinewidth=2pt,
	frametitlerule=false,
	backgroundcolor=black!5!white,
	frametitle={Command Line},
	frametitlefont={\normalfont\sffamily\color{white}\hspace{-1em}},
	frametitlebackgroundcolor=black!50!white,
	nobreak,
}

\mdfdefinestyle{file}{
	innertopmargin=1.6\baselineskip,
	innerbottommargin=0.8\baselineskip,
	topline=false, bottomline=false,
	leftline=false, rightline=false,
	leftmargin=2cm,
	rightmargin=2cm,
	singleextra={%
		\draw[fill=black!10!white](P)++(0,-1.2em)rectangle(P-|O);
		\node[anchor=north west]
		at(P-|O){\ttfamily\mdfilename};
		\def\l{3em}
		\draw(O-|P)++(-\l,0)--++(\l,\l)--(P)--(P-|O)--(O)--cycle;
		\draw(O-|P)++(-\l,0)--++(0,\l)--++(\l,0);
	},
	nobreak,
}

\mdfdefinestyle{question}{
	innertopmargin=1.2\baselineskip,
	innerbottommargin=0.8\baselineskip,
	roundcorner=5pt,
	nobreak,
	singleextra={%
		\draw(P-|O)node[xshift=1em,anchor=west,fill=white,draw,rounded corners=5pt]{%
			Question \theQuestion\questionTitle};
	},
}

\newcounter{Question} 

\mdfdefinestyle{warning}{
	topline=false, bottomline=false,
	leftline=false, rightline=false,
	nobreak,
	singleextra={%
		\draw(P-|O)++(-0.5em,0)node(tmp1){};
		\draw(P-|O)++(0.5em,0)node(tmp2){};
		\fill[black,rotate around={45:(P-|O)}](tmp1)rectangle(tmp2);
		\node at(P-|O){\color{white}\scriptsize\bf !};
		\draw[very thick](P-|O)++(0,-1em)--(O);
	}
}

\mdfdefinestyle{info}{%
	topline=false, bottomline=false,
	leftline=false, rightline=false,
	nobreak,
	singleextra={%
		\fill[black](P-|O)circle[radius=0.4em];
		\node at(P-|O){\color{white}\scriptsize\bf i};
		\draw[very thick](P-|O)++(0,-0.8em)--(O);
	}
}

\usepackage{setspace}

\usepackage{graphicx}
\usepackage{amssymb}
\usepackage{multirow}

\usepackage{verbatim}

\usepackage{amsmath}
\usepackage{authblk}

\usepackage{xr}

\makeatletter
\newcommand*{\addFileDependency}[1]{
  \typeout{(#1)}
  \@addtofilelist{#1}
  \IfFileExists{#1}{}{\typeout{No file #1.}}
}
\makeatother
 
\newcommand*{\myexternaldocument}[1]{%
    \externaldocument{#1}%
    \addFileDependency{#1.tex}%
    \addFileDependency{#1.aux}%
}
\myexternaldocument{SuppMaterials}
 
\begin{document}




\title{Weighted graphlets and deep neural networks for protein structure classification}


\renewcommand\Authfont{\fontsize{12}{14.4}\selectfont}
\renewcommand\Affilfont{\fontsize{9}{10.8}\itshape}
\author[1]{Hongyu Guo}
\author[2,3,4]{Khalique Newaz}
\author[5]{Scott Emrich}
\author[2,3,4]{Tijana Milenkovi\'c}
\author[1,*]{Jun Li}
\affil[1]{Department of Applied and Computational Mathematics and Statistics, University of Notre Dame, Notre Dame, IN 46556, USA}
\affil[2]{Department of Computer Science and Engineering, University of Notre Dame, Notre Dame, IN 46556, USA}
\affil[3]{Eck Institute for Global Health, University of Notre Dame, Notre Dame, IN 46556, USA}
\affil[4]{Interdisciplinary Center for Network Science \& Applications, University of Notre Dame, Notre Dame, IN 46556, USA}
\affil[5]{Department of Electrical Engineering \& Computer Science, University of Tennessee at Knoxville, TN 37996, USA}
\affil[*]{To whom correspondence should be addressed. Tel: +1 574 631 3429; Fax: +1 574 631 4822; Email: jun.li@nd.edu}

\date{}
\maketitle


\doublespacing

\begin{abstract}
As proteins with similar structures often have similar functions, analysis of protein structures can help predict protein functions and is thus important. We consider the problem of protein structure classification, which computationally classifies the structures of proteins into pre-defined groups. We develop a weighted network that depicts the protein structures, and more importantly, we propose the first graphlet-based measure that applies to weighted networks. Further, we develop a deep neural network (DNN) composed of both convolutional and recurrent layers to use this measure for classification. Put together, our approach shows dramatic improvements in performance over existing graphlet-based approaches on 36 real datasets. Even comparing with the state-of-the-art approach, it almost halves the classification error. In addition to protein structure networks, our weighted-graphlet measure and DNN classifier can potentially be applied to classification of other weighted networks in computational biology as well as in other domains.
\end{abstract}

Proteins are the building molecules of life, and their diverse functions define the mechanisms of sophisticated organisms \cite{alberts2015molecular}. Wet-lab experiments can help determine functions of proteins, but they are expensive and time-consuming \cite{lee2007predicting, slabinski2007challenge}. As economical alternatives, computational methods have been proposed. Previous studies have shown that functions of a protein can be inferred from other proteins with similar structures \cite{mills2015biochemical}, and consequently, protein structure analysis is essential for the computational prediction of protein functions. In this paper, we focus on ``protein structure classification,'' also known as ``protein structural classification,'' a supervised problem of assigning structures of proteins into pre-defined classes \cite{kasabov2013springer}. 

A method for protein structure classification typically consists of two steps: feature extraction and classifier training. In the first step, data for each protein (such as the sequence or 3D coordinates of the given protein's amino acids) is summarized into a ``measure,'' which is a set of attributes/features.  Like in other applications of classification, a measure can be a vector of attributes, such as the weight and the height of a person, or a matrix of attributes, such as the pixel values in a 2D image. In the second step, this measure is used as the input to train a classifier.

Many measures have been developed for extracting features for protein structure classification, and they can be roughly divided into two categories: sequence-based and 3D-structure-based. Sequence-based measures extract information from the amino acid sequences of proteins \cite{marks2012protein, dill2012protein, yang2016sixty}. Examples include a simple measure that describes the frequencies of the different amino acids \cite{faisal2017grafene}, and more complicated ones that also consider sequence positions of individual amino acids \cite{altschul1997gapped, jones1999protein, remmert2012hhblits, xia2016ensemble}. However, cases have been found that proteins with very similar sequences have distinct 3D structures, on which sequence-based measures struggle \cite{pearson2005limits, krissinel2007relationship, kosloff2008sequence}. 3D-structure-based measures, instead, extract information from 3D coordinates of amino acids in proteins. These coordinates are typically available as Protein Data Bank (PDB) files. Some early measures in this category derive protein structure features directly from the 3D coordinates, for example, by subdividing the bounding box of a protein and examining the distribution of certain atoms across these subsections \cite{cui2008classification}, or by applying Spherical Trace Transformation to the 3D coordinates and generating structure features by evaluating the coefficients of the transformation \cite{ kalajdziski2007protein}. Recently, network-based measures, which are another type of 3D-structure-based measures, have become more popular: given a protein, the 3D coordinates of its amino acids \cite{berman2000protein} are first used to build a protein structure network (PSN) with nodes representing amino acids and edges representing spatial closeness between amino acids, and then various measures are derived from the PSN topology, including global, high-level descriptive statistics \cite{faisal2017grafene} (e.g., average degree, average closeness centrality, and average clustering coefficient) and local, more detailed graphlet-based measures \cite{faisal2017grafene, newaz2018network}. In this paper, we focus on graphlet-based measures, as previous studies \cite{faisal2017grafene, newaz2018network} have demonstrated their higher accuracy for protein structure classification than the other (i.e., non-graphlet) network-based measures, as well as direct-3D-structural-based and sequence-based measures.

\begin{figure}[h]
	\centering\includegraphics[width=1\linewidth]{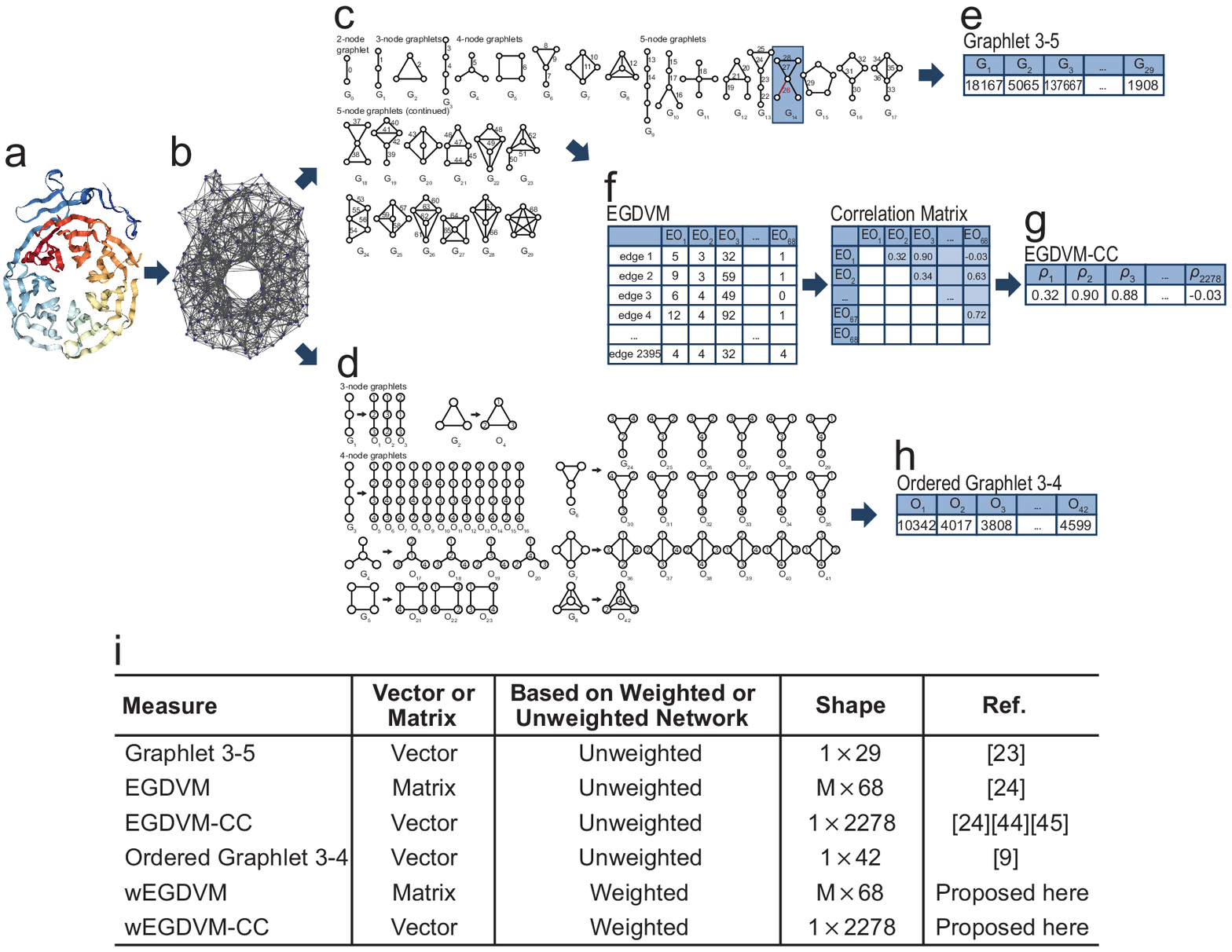}
	\caption{Existing graphlet-based measures. \textbf{(a)} The 3D structure of protein domain 1ERJ (PDB ID) \cite{sprague2000structure} generated by the NGL viewer \cite{rose2018ngl} from RCSB PDB (www.rcsb.org) \cite{berman2000protein}; \textbf{(b)} An unweighted protein structure network derived from the PDB file; \textbf{(c)} 29 graphlets and 68 graphlet edge orbits (numbered on the edges of graphlets) defined on up to 5 nodes. The highlighted graphlet edge orbits EO$_{26}$ will be used to illustrate the definition of wEGDVM in Fig. \ref{fig:wEGDVM}; \textbf{(d)} 42 ordered graphlets defined on up to 4 nodes; \textbf{(e)} Graphlet-3-5, a length-29 vector; \textbf{(f)} EGDVM, a matrix with 68 columns and 2395 (different from protein to protein) rows; \textbf{(g)} EGDVM-CC, a length-2278 vector by using the upper-diagonal elements of the correlation matrix of EGDVM; \textbf{(h)} Ordered graphlet 3-4, a length-42 vector; \textbf{(i)} Main characteristics of graphlet-based measures. In the ``Shape'' column, $M$ is the number of edges in a graph.}
	\label{fig:graphlet}
\end{figure}

So far, two graphlet-based measures of network topology have been used for protein structure classification, one based on (the original) graphlets \cite{prvzulj2004modeling}, and the other based on ordered graphlets \cite{malod2014gr, faisal2017grafene}. Fig. \ref{fig:graphlet} illustrates how these measures are defined/calculated. Graphlets are small connected non-isomorphic induced subgraphs of a network (Fig. \ref{fig:graphlet}c), and ordered graphlets are a refined version of graphlets that consider the order of the nodes (Fig. \ref{fig:graphlet}d; in protein structure classification, the node order is the relative order of amino acids in the protein sequence.). The measure based on the original graphlets, named Graphlet-3-5, summarizes the network structure by counting the occurrences of each of the 29 graphlets defined with up to five nodes in the network (Fig. \ref{fig:graphlet}e). It is a vector of counts, with length 29 (Fig. \ref{fig:graphlet}i). Similarly, the measure based on ordered graphlets, named Ordered-Graphlet-3-4, is a vector of the occurrences of each of the 42 ordered graphlets defined with up to four nodes in the network (Fig. \ref{fig:graphlet}h), which is of length 42 (Fig. \ref{fig:graphlet}i). 

Both of the above two graphlet-based measures are in the form of vectors. Using these vector-form measures for classification is straightforward: traditional classifiers such as logistic regression and support vector machines take vector-form measures as the input, and thus researchers typically pick up an off-the-shelf classifier and use it with such vector-form measures. In this case, there is no apparent need to develop a specific classifier tailored for these measures. 

However, in addition to these two measures, there exists a powerful graphlet-based measure called the edge graphlet degree vector matrix (EGDVM, \cite{solava2012graphlet}) that was proposed in the task of clustering edges in a protein-protein interaction network but can potentially be used for protein structure classification. EGDVM is based on graphlet edge orbits, which are unique positions of edges in graphlets (Fig. \ref{fig:graphlet}c). Different from Graphlet-3-5 and Ordered-Graphlet-3-4, which are vectors that only record the occurrences of (the original or ordered) graphlets in the whole network, EGDVM is a matrix that records the occurrences of graphlets around \textit{every} edge. The $(i, j)^{th}$ element of EGDVM is the count of how many times edge $i$ touches graphlet edge orbit $j$. The number of columns of EGDVM equals the number of graphlet edge orbits, which is 68 for graphlets on up to five nodes, and the number of rows equals the number of edges of the network (Fig. \ref{fig:graphlet}i; in our task, a network would be a PSN). Note that PSNs for different proteins have different numbers of edges, and thus the number of rows of EGDVM is different from protein to protein. We call a matrix like this ``matrix of variable size'' or ``variable-size matrix'' in this paper.

EGDVM could be more informative than the other two measures, as it keeps the detailed graphlet configurations for every single edge and thus is able to describe local diversities of the network topology. Unfortunately, as a matrix with variable size, it cannot be used as the input by any off-the-shelf classifier, which can only take a vector of constant length as the input. As a result, EGDVM has never been used for protein structure classification. Here we propose a naive remedy: converting EGDVM into a fixed length vector, and then apply an off-the-shelf classifier. This conversion can be done by following the idea from two previous studies \cite{yaverouglu2014revealing, trpevski2016graphlet}, which proposed methods for converting matrix measures (not EGDVM) to vector measures: first computing the Pearson's correlation matrix (Fig. \ref{fig:graphlet}f) of EGDVM, whose elements are the Pearson's correlation coefficients of pairs of columns of EGDVM, and then concatenating the upper diagonal elements of the correlation matrix into a vector (Fig. \ref{fig:graphlet}g). We call this vector EGDVM-CC (CC stands for correlation coefficient), which is of length $68 \times 67 / 2 = 2278$. We admit that such a conversion is far from satisfactory: Pearson's correlation coefficients only capture globally linear relationships between columns of EGDVM, and thus the information in EGDVM about local diversities of the network topology is likely lost during the conversion.

We have described three graphlet-based measures: Graphlet-3-5, Ordered-Graphlet-3-4, and EGDVM-CC. The first two are the only graphlet-based measures that have been used for protein structure classification to date, while the last one is converted from EGDVM, a matrix measure of variable size that cannot be used by off-the-shelf classifiers. Although EGDVM-CC is proposed by us in this paper, to differentiate from the novel measure we propose later in this paper, we still call EGDVM-CC, together with Graphlet-3-5 and Ordered-Graphlet-3-4, ``the three existing measures''.  All these three measures are vectors of constant length. A more detailed review of these graphlet-based measures is given in Supplementary Materials, and the main characteristics of these measures are summarized in Fig. \ref{fig:graphlet}i.

A common limitation of the three existing measures, as well as EGDVM, is that they are all based on unweighted PSNs. Actually, no graphlet-based measures, including but not limited to the several measures discussed above, have ever been developed for weighted networks. However, weighted networks are widely used in various applications, and they can be more suitable/informative than unweighted networks. Take PSN as an example. An unweighted network connects a pair of nodes by an edge if the spacial distance of the two amino acids is less than a threshold, but it does not keep the information of how close in space the two amino acids are: they can be right next to each other or barely below the cutoff. This information can be retained by weighted networks, which add a weight to each edge. In this paper, we introduce the concept of a weighted PSN with weights depending on both distances in space and distances in sequence. These weights effectively reflect the importance of edges to the protein structure classification.

\begin{figure}[h]
	\centering
	\includegraphics[width=1\linewidth]{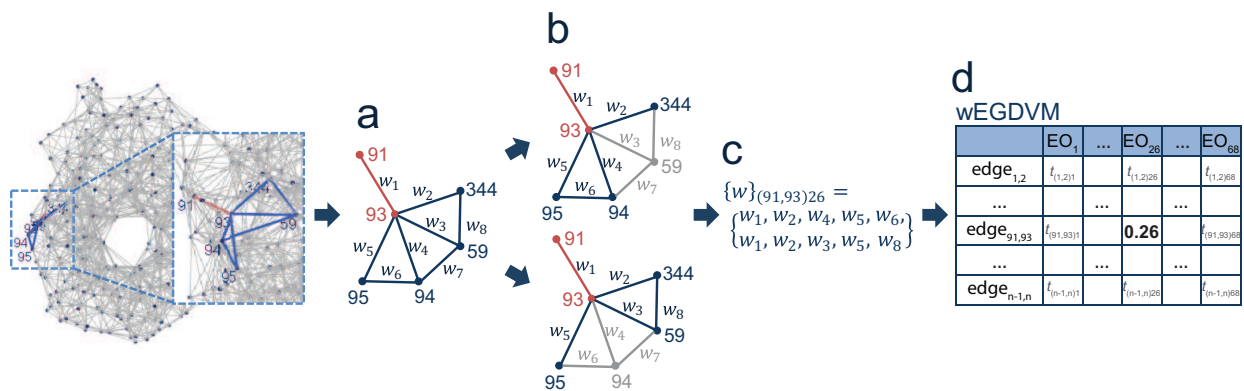}
	\caption{Illustration of wEGDVM applied to protein domain 1ERJ (chain A). This protein domain has 350 amino acids.  The corresponding weighted PSN has 2395 edges when cutoff 6\AA{} for the space distance is used. wEGDVM of protein 1ERJ is thus of size $2395 \times 68$. For the purpose of simplification, we focus on the calculation of statistic $t_{(91,93)26}$ corresponding to edge $e_{91,93}$ and graphlet edge orbit $EO_{26}$. The rest $t_{(ij)l}$'s in wEGDVM are calculated in the same manner. \textbf{(a)} Edge $e_{91,93}$ touches graphlet edge orbit $EO_{26}$ twice, which involves nodes $\{v_{59}, v_{91}, v_{93}, v_{94}, v_{95}, v_{344}\}$. \textbf{(b)} The first graphlet edge orbit $EO_{26}$ is formed by nodes $\{v_{91}, v_{93}, v_{94}, v_{95}, v_{344}\}$ (upper). The second graphlet edge orbit $EO_{26}$ is formed by nodes $\{v_{59}, v_{91}, v_{93}, v_{95}, v_{344}\}$ (lower). \textbf{(c)} The weights of graphlets corresponding to the graphlet edge orbit touched are summarized into the multiset $\{w\}_{(91,93)26}$. \textbf{(d)} The Cram\'er-von-Mises statistic $t_{(91,93)26} = 0.26$ is calculated based on $\{w\}_{(91,93)26}$ and all weights.}
	\label{fig:wEGDVM}
\end{figure}

More importantly, we propose the first graphlet-based measure that applies to weighted networks. This measure, named wEGDVM (weighted EGDVM), is a matrix measure of the same dimension as EGDVM (an example is shown in Fig. \ref{fig:wEGDVM}), and thus it contains information about local diversities of the network topology, but the elements of wEGDVM are defined in a novel way compared to those of EGDVM so that they can efficiently utilize the edge weights in the weighted network.

\begin{figure}[h]
	\centering\includegraphics[width=1\linewidth]{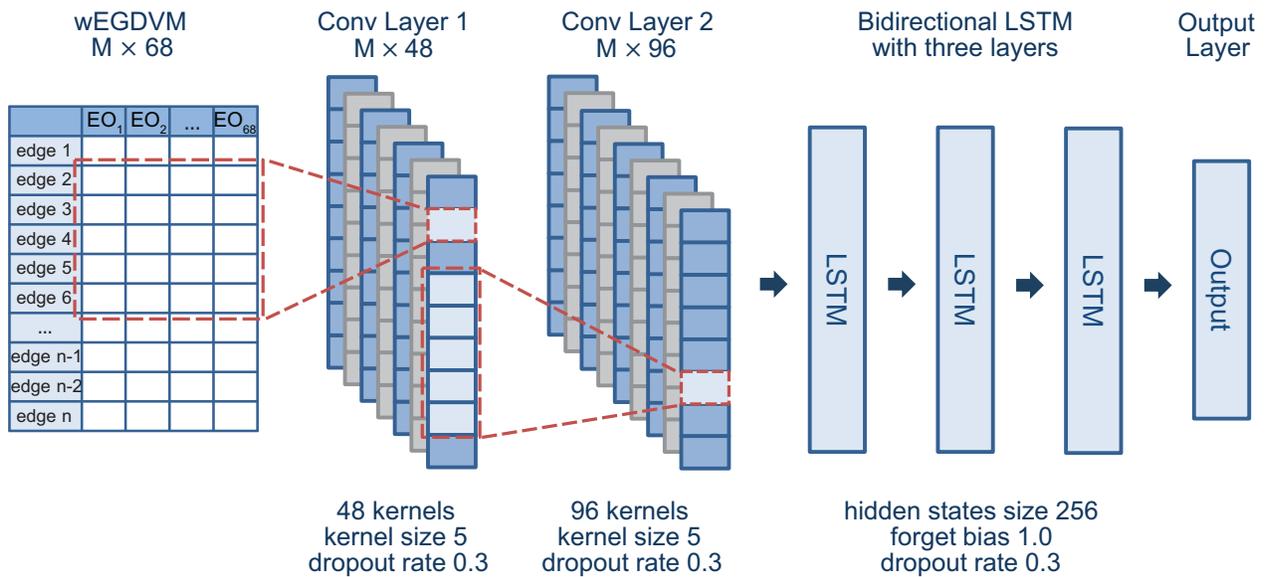}
	\caption{The structure of the proposed DNN classifier. The input of this network is wEGDVM of variable size $M \times 68$, which is first processed by two convolutional layers. 48 kernels are assigned to the first convolutional layer. 96 kernels are assigned to the second convolutional layer. Zero padding is applied to keep the height of each convolutional layer $M$. The output of the second convolutional layer is of variable size $M \times 96$ , which is further processed by a bidirectional LSTM network with three layers and hidden state size 256. The outputs of LSTM are summed up to a fixed length vector of size 512, which is fully connected to an output layer.}
	\label{fig:DNN}
\end{figure}

However, wEGDVM, the measure we propose, just like the existing EGDVM measure, cannot be used as the input of any off-the-shelf classifier as it is a matrix of variable size. To overcome this difficulty, we develop a new classifier that takes the whole matrix measure as the input without any transformation or concatenation. The approach is based on deep neural networks (DNNs), which have achieved enormous success on complex tasks in fields like computer vision \cite{voulodimos2018deep} and natural language processing \cite{young2018recent}. Our DNN, as illustrated in Fig. \ref{fig:DNN}, consists of several layers of convolutional neural networks (CNNs) followed by several layers of recurrent neural networks (RNNs). Such a structure has been shown as an efficient design to extract patterns from sequential data \cite{zhou2015c, jongejan2016, ullah2018action}, and we think that it also fits our needs very well: CNNs may capture local protein structures such as turns in $\alpha$-helices and strands in $\beta$-sheets, and RNNs may subsequently integrate these local structures into larger components such as $\alpha$-helices and $\beta$-sheets and finally give an overall impression of the structure of the protein.

We test our new framework for protein structure classification, which includes a weighted network, a new graphlet-based matrix measure to summarize the weighted network topology, and a DNN-based classifier, on 36 real protein domain datasets of very different sample sizes and numbers of classes. Our new framework shows dramatic improvements in classification accuracy over methods based on the three existing graphlet-based measures, including the measure based on ordered graphlets, the current state-of-the-art.

\section*{Real datasets}
To evaluate the proposed method, we use datasets from a previous large-scale protein structure classification study \cite{newaz2018network}, including 36 datasets with 9,440 protein domains annotated by the CATH database \cite{orengo1997cath} and 13,068 protein domains annotated by the SCOP database \cite{murzin1995scop}. Protein domains are compact continuous structural regions of proteins that may exist, fold, and function independently \cite{wetlaufer1973nucleation, bork1991shuffled}. Following previous studies of protein structure classification, we use protein domains instead of proteins as samples, and we use the two phrases, ``protein domains'' and ``proteins,'' interchangeably. The sample sizes (numbers of proteins) of these datasets range from 69 to 11,362, and the numbers of classes range from two to 33. Details are shown in Fig. \ref{fig:samplesize}. The detailed procedure for collecting these datasets is described in the Supplementary Materials.

\section*{Design of comparisons}
Our new method for protein structure classification consists of three novel sessions: a weighted network for depicting protein structures, a matrix measure wEGDVM for the weighted network, and a DNN classifier that takes wEGDVM as the input. Since our method differs from existing approaches in both measures and classifiers, instead of simply studying whether it as a whole works better than existing approaches, we break the aim down and test two hypotheses: (1) whether wEGDVM captures more information about protein structures than existing measures, and (2) whether our DNN classifier is able to efficiently utilize such information. 

Testing the first hypothesis requires using the same classifier for different measures, so that the difference in performance due to different classifiers is erased. However, Graphlet-3-5, EGDVM-CC (EGDVM is not considered as no existing classifiers have been developed for it), and Ordered-Graphlet-3-4 are vectors of constant length and use off-the-shelf-classifiers, while wEGDVM is a matrix of variable size and uses our DNN classifier. To make the comparison possible, we convert wEGDVM into a vector, which we call wEGDVM-CC, following the same procedure as converting EGDVM to EGDVM-CC: first computing its Pearson's correlation matrix and then concatenating the upper diagonal elements of the correlation matrix. Then we apply ($\ell_2$ regularized) logistic regression, an off-the-shelf classifier, on wEGDVM-CC, as well as Graphlet-3-5, EGDVM-CC, and Ordered-Graphlet-3-4. Note that such a comparison only tells how informative wEGDVM-CC is (compared to other measures), not wEGDVM. wEGDVM-CC is only a degenerated version of wEGDVM and is less informative as during conversion it may have lost most information about local diversities of the network topology contained in wEGDVM. Thus, if wEGDVM-CC is more informative than another measure, wEGDVM should also be, but if wEGDVM-CC is not more informative than another measure, wEGDVM may still be.

To test the second hypothesis, we apply our DNN classifier directly to wEGDVM without any conversion and compare this approach with the approach of applying logistic regression to wEGDVM-CC as well as other measures. This is the best that we can do for the comparison, since we cannot run the DNN classifier on wEGDVM-CC nor logistic regression on wEGDVM. The latter is exactly the reason why we needed to develop the DNN classifier in the first place. 

The criterion we use for comparison is the misclassification rate: the proportion of samples (protein domains) that are incorrectly classified. A lower misclassification rate corresponds to a higher accuracy and is thus favored. The misclassification rate is calculated based on stratified 5-fold cross-validation (CV). For short, we use ``CV error'' for this ``5-fold CV misclassification rate'' hereafter.

\section*{Comparison of different measures}
To study the ability of different network measures to capture structural information, we apply logistic regression on four measures: Graphlet-3-5, EGDVM-CC, Ordered-Graphlet-3-4, and wEGDVM-CC. Among them,  EGDVM-CC and wEGDVM-CC can be viewed as degenerated versions of their corresponding matrix measures to bypass the inability of logistic regression in handling variable-size matrices. We chose logistic regression of all off-the-shelf classifiers as it has been shown to work the best among off-the-shelf classifiers for existing measures \cite{newaz2018network}, in the sense of both high classification accuracy and high computational efficiency. The CV errors of different methods on the 36 datasets are shown in Fig. \ref{fig:performance}. 

We first check the results for the three existing methods: Graphlet-3-5, EGDVM-CC, and Ordered-Graphlet-3-4. It is clear that consistently across datasets, Ordered-Graphlet-3-4 outperforms EGDVM-CC, which in turn outperforms Graphlet-3-5. These results can be helpful in understanding the three measures' different ways of summarizing topological information of the network. EGDVM-CC, despite losing much of the information during the conversion from EGDVM, is still more informative than Graphlet-3-5. Ordered-Graphlet-3-4 is even more informative than EGDVM-CC, despite it only considering subgraphs up to four nodes instead of five, suggesting the importance of including the order information. However, although Ordered-Graphlet-3-4 outperforms EGDVM-CC, it does not necessarily mean that including the order information is more important than including information about local diversities of the network topology, as EGDVM-CC loses most of such information contained in EGDVM during the conversion. Unfortunately, as no current classification methods are able to handle EGDVM without conversion, a direct comparison between EGDVM and Ordered-Graphlet-3-4 is not possible.

Next, we compare wEGDVM-CC with the three existing methods. Just like Ordered-Graphlet-3-4, wEGDVM-CC outperforms Graphlet-3-5 by a large margin (average CV error 0.118 vs 0.253, a 47\% reduction). Most importantly, it also significantly outperforms EGDVM-CC, its unweighted predecessor: wEGDVM-CC achieves lower CV errors on 34 out of 36 datasets, and only performs slightly worse on the other two (differences in CV error 0.013 and 0.006, respectively). On average, compared to EGDVM-CC, the CV error of wEGDVM-CC decreases by 39\%. And on dataset scop-second-alphabeta, where EGDVM-CC gives the highest CV error among all the 36 datasets,wEGDVM-CC reduces the CV error by almost a half (0.225 vs 0.434). These results clearly suggest that wEGDVM-CC extracts more topological information from the network than EGDVM-CC, indicating the success of our approaches in assigning proper weights to the network edges and in choosing proper statistics that summarize the weights on graphlets.

The performance of wEGDVM-CC is only slightly worse than that of Ordered-Graphlet-3-4, the current state-of-the-art. Although wEGDVM-CC shows inferior performance on 25 out of 36 datasets, the differences are not pronounced (average CV error difference 0.022).

It is worth noting again that wEGDVM-CC is only a degenerated version of wEGDVM, our proposed measure for weighted graphlet topology. The inferior performance of wEGDVM-CC to Ordered-Graphlet-3-4 does not mean that wEGDVM is less informative than Ordered-Graphlet-3-4. The construction of wEGDVM-CC is only for the comparison in this section; we do not recommend using it otherwise. In the next section, we will show that wEGDVM, with the help from DNNs, shows highly boosted performance and beats Ordered-Graphlet-3-4 by a large margin. 

\section*{Evaluation of the DNN classifier}
\begin{figure}[h]
	\centering\includegraphics[width=1\linewidth]{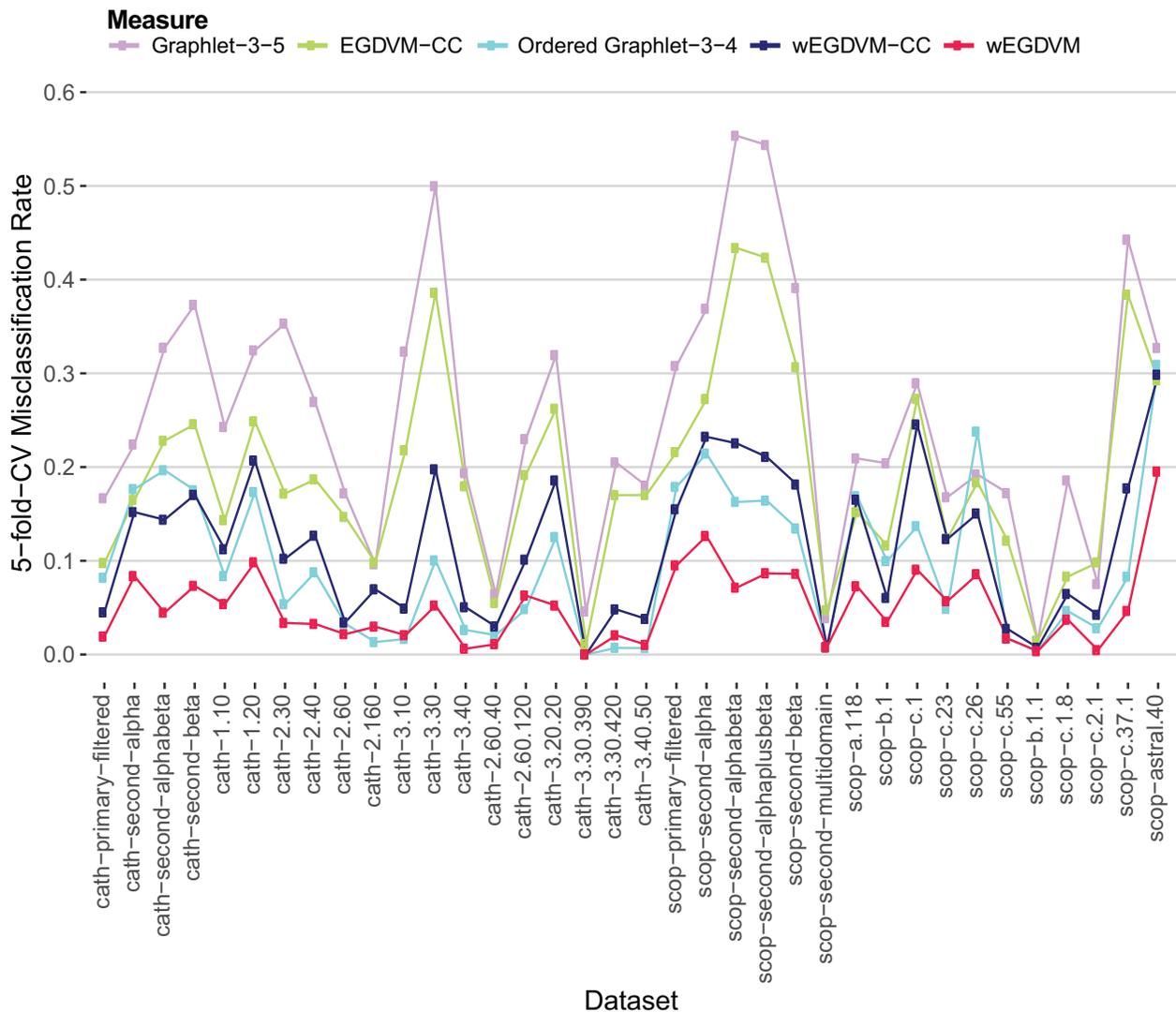}
	\caption{CV errors of different graphlet-based approaches on the 36 protein domain datasets. Five different approaches are considered. They are shown in different colors, and the color code is shown on the top. The DNN classifier is used for wEGDVM, while logistic regression is used for the other four measures.}
	\label{fig:performance}
\end{figure}

Next, we apply our DNN classifier to the matrix-form wEGDVM to see whether this approach elevates the performance of the approach of applying logistic regression to wEGDVM-CC. For short, we call these two approaches ``wEGDVM + DNN'' and ``wEGDVM-CC + logistic regression'' in this section. The CV errors on the 36 datasets are again shown in Fig. \ref{fig:performance}. 

The elevation in performance is dramatic. wEGDVM + DNN outperforms wEGDVM-CC + logistic regression on 34 out of the 36 datasets and ties on the other two, on which the CV errors of wEGDVM-CC + logistic regression are already very low (0 and 0.0077), leaving little room for improvements. On average over the 36 datasets, wEGDVM + DNN reduces the CV error by more than a half (0.051 vs 0.118, a 57\% reduction). This significant improvement in performance indicates that wEGDVM contains considerably more information about network topology compared to its degenerated vector form, and that the DNN classifier efficiently utilizes this information.

wEGDVM + DNN even outperforms Ordered-Graphlet-3-4 by a large margin, reducing the average CV error by almost a half (0.051 vs 0.096, a 47\% reduction). wEGDVM + DNN gives lower CV errors on 26 out of 36 datasets. On the 10 datasets that wEGDVM + DNN does not outperform (including three ties), the CV errors of both methods are low and the differences between the two methods are small (average CV error 0.017 for Ordered-Graphlet-3-4 and 0.023 for wEGDVM). On the contrary, on the 15 datasets where Ordered-Graphlet-3-4 has CV errors higher than 0.10, wEGDVM + DNN cuts the CV errors by 33\% to 77\% (51\% on average). What is especially interesting is the scop-astral.40 dataset (the last column in Fig. \ref{fig:performance}): while none of Ordered-Graphlet 3-4, EGDVM-CC, and wEGDVM-CC give noticeable improvements over Graphlet-3-5, wEGDVM + DNN reduces the CV error from around 0.30 to below 0.20.

Combing the results in the previous section and this section is very interesting. Under logistic regression, wEGDVM has to be converted to wEGDVM-CC, which does not compete with Ordered-Graphlet-3-4. However, with the help of the DNN classifier, wEGDVM beats Ordered-Graphlet-3-4 by a large margin. This clearly proves the need and power of the DNN classifier, and also shows that wEGDVM is a much more informative measure than Ordered-Graphlet-3-4 as well as the other existing graphlet-based measures.

\section*{Conclusions and discussion}
\label{S:5}
For classifying proteins into pre-specified classes, we have presented a new way to build weighted protein structure networks, where weights are defined in a way to highlight informative relationships between amino acids. More importantly, we have also developed a novel matrix measure that retains the graphlet configurations of individual edges, and the elements of the matrix are based on statistics that describe the distributions of the weights. This is the first graphlet-based measure that is designed for weighted networks. Results on 36 real datasets show that this new measure of weighted network topology contains much more information about protein structures than existing approaches. Further, we have developed a DNN classifier that directly processes the variable-size matrix measure that we propose. Overall, our pipeline dramatically improves the performance over all existing graphlet-based methods.

wEGDVM and the DNN classifier that we have developed in this paper can be used for applications other than protein structure classification. Generally, given a weighted network, wEGDVM can always be calculated and used as a valid measure of its topology. Moreover, since the Cram\'er-von-Mises statistic that wEGDVM uses (see Methods) is nonparametric, wEGDVM is invariant under any monotone transformation of the weights, and this invariance further facilitates applying it to networks constructed from data from other realms. Once wEGDVM is constructed, our DNN classifier should be applicable to it. Further, the DNN classifier may also be applied to matrix measures other than wEGDVM, such as matrix measures of which the elements are not defined by the Cram\'er-von-Mises statistic.

The Cram\'er-von-Mises statistic in wEGDVM can be replaced by other statistics, for example, the Kolmogorov-Smirnov (KS) statistic \cite{smirnov1939estimate} or the Kuiper statistic \cite{kuiper1960tests}.  These two statistics show inferior performance to the Cram\'er-von-Mises statistic on our data (detailed results not shown).

Literature has shown that the performance of a DNN can highly depend on the structure and other hyperparameters of the network. In this study, we choose a small set of candidates (values given in the Supplementary Materials) for each hyperparameter and train our model under all combinations of these hyperparameters. We conduct the search of hyperparameters on three datasets with relatively small sample sizes (cath-1.20, cath-3.30.420, and scop-c.1.8) and identify a combination of hyperparameters (described in the Supplementary Materials) that works well on all the three datasets. Then we utilize this combination on all the 36 datasets we use. While an alternative approach that searches for a dataset-specific optimal combination of hyperparameters for every dataset may lead to further improvements in misclassification rates, the current excellent performance under the common set of hyperparameters for all datasets could suggest that the DNN classifier we propose is not too sensitive to the choice of hyperparameters and that the current values of hyperparameters we use may be applicable to a wide range of data.

\pagebreak


\section*{Methods}
We introduce our new approach in the following three sections. First, we describe how we construct a weighted PSN, especially, how we assign edge weights. Second, we describe our statistical approach for obtaining wEGDVM from the weighted PSN. Third, we introduce our DNN classifier that processes wEGDVM directly.

\subsection*{Weighted PSNs}\label{subsec:meth1}
To build a weighted network, we first build a network with the same frame (nodes and edges) as the unweighted network proposed in \cite{faisal2017grafene} and then add weights to its edges. Given a protein with $n$ amino acids, the nodes $\{v_i\}_{i = 1,...,n}$ of the network represent the amino acids, and the edges $\{ e_{ij} \}_{i, j = 1,...,n}$ represent connections between the nodes, where the indexes $i$ and $j$ are the sequence positions of the nodes.  Let $r_{ik_i} = (x_{ik_i}, y_{ik_i}, z_{ik_i})$ be the 3D coordinates of heavy atom $k_i$ in amino acid $i$. 
Faisal \textit{et al}. \cite{faisal2017grafene} defined the space distance $d^{space}_{ij}$ between a pair of amino acids as the closest space distance between any of their heavy atoms
\begin{equation}
\label{eq:spacedist}
d^{space}_{ij} = \min_{k_i,k_j} \|r_{ik_i}-r_{jk_j}\|.
\end{equation}
Given a space distance cutoff $c$, an edge $e_{ij}$ between amino acid pairs $(v_i, v_j)$ exists if and only if $d^{space}_{ij} < c$. A properly selected $c$, which determines the number of edges, was shown to be important, as an overly large $c$ leads to a random-like network structure while an overly small $c$ leads to a disconnected network structure \cite{milenkovic2009optimized}. We try $c=4$\AA{} and $c=6$\AA{}, as suggested in \cite{milenkovic2009optimized}, and find 6\AA{} to be a better choice (details given in Supplementary Materials and Fig. \ref{fig:cutoff}), which is thus used for all of our computations.

Now we consider how to add weights to the edges. Intuitively, a pair of amino acids that are close to each other in space should have a large weight, and thus a straightforward idea is to use a weight determined by the space distance, such as $w_{ij}=1/d^{space}_{ij}$ used in previous studies \cite{chakrabarty2016naps}. This is a reasonable choice; however, since a pair of amino acids that are very close to each other in sequence (such as amino acids next to each other) are naturally also close in space, the edge between them will always have a large weight, when this definition of weights is used. However, this large weight is not related to the structure of the network and is thus not informative. To down-weight this edge, we define the weight $w_{ij}$ for edge $e_{ij}$ as
\begin{equation}
\label{eq:weightsdef}
w_{ij} = \sqrt{\frac{d^{sequence}_{ij}}{d^{space}_{ij}}},
\end{equation} 
where $d^{sequence}_{ij}$ is the distance in sequence
\begin{equation}
\label{eq:sequencedist}
d^{sequence}_{ij} = |i-j|.
\end{equation}
Our definition gives the largest weights to amino acids that are far away in sequence but are close by in space, which is arguably the most informative relationship in protein structures, as it indicates a folding. The square root is used to stabilize the variance.

\subsection*{Weighted edge graphlet degree vector matrix}\label{subsec:meth2}
All of the three existing graphlet-based measures (Graphlet-3-5, EGDVM/EGDVM-CC, and Ordered-Graphlet-3-4) are based on unweighted networks.  In this section, we introduce the first graphlet-based measure designed for weighted networks: wEGDVM. Like EGDVM, wEGDVM is also a matrix measure of size $M\times 68$, where $M$ is the number of edges in the network, but elements of this matrix are defined in a different way.

Given an edge $e_{ij}$ and a graphlet edge orbit $EO_l$, we first collect the edge weights of all graphlets corresponding to $EO_l$ that are touched by $e_{ij}$. These weights form a multiset $\{w\}_{(ij)l}$. Then we summarize this multiset into a single statistic and use this statistic as an element of wEGDVM.

An intuitive choice of the statistic is the sum of all elements in $\{w\}_{(ij)l}$. It is easy to show that under this choice, wEGDVM coincides with EGDVM on unweighted networks, which can be viewed as weighted networks with constant weights 1 for all edges. However, such a definition may not be a good choice. For example, ten small weights (denoting ten not-so-close pairs) have different implications for protein structures compared to a single large weight (denoting a close-by pair) even if the sum of the ten small weights is the same as the large weight. The more important information about protein structures seems to lie on the statistical distribution of weights.

The statistic we decide to use is the Cram\'er-von-Mises statistic \cite{anderson1962distribution} that measures the deviance of the distribution of $\{w\}_{(ij)l}$ from the distribution of $\{w\}_{PSN}$, the set of all weights of the PSN. Assume the size of $\{w\}_{(ij)l}$ is $M_l$ and the size of $\{w\}_{PSN}$ is $M$, and let $r_p$ and $s_q$ be the ranks in the pooled set of the ordered observations of the first and the second weight sets. The Cram\'er-von-Mises statistic $t_{(ij)l}$ is
\[
t_{(ij)l} = \frac{U}{M_lM(M_l+M)}-\frac{4M_l M-1}{6(M_l+M)},
\]
where
\[
U = M_l\sum^{M_l}_{p = 1}(r_p-p)^2 + M\sum^{M}_{q = 1}(s_q-q)^2.
\]

We use $t_{(ij)l}$ of different $(ij)$'s and $l$'s to fill the wEGDVM. We define the weighted edge graphlet degree vector (wEGDV) for edge $e_{ij}$ as $[t_{(ij)l}]_{l = 1,...,68}$, and then construct wEGDVM by combining wEGDVs of all edges as rows. The rows in wEGDVM are ordered hierarchically, first by edge index $i$ and then by edge index $j$. For a weighted PSN with $M$ edges, the matrix measure wEGDVM consists of $M$ rows and 68 columns (Fig. \ref{fig:graphlet}i). Again, this is a variable-size matrix. The procedure of deriving wEGDVM for protein domain 1ERJ (chain A) is illustrated in Fig. \ref{fig:wEGDVM} as an example. 

\subsection*{Deep neural network based classifier}\label{subsec:meth3}
Converting EGDVM to EGDVM-CC loses its main advantage of describing local diversities of the network topology. To avoid such loss for wEGDVM, we train a DNN classifier that directly handles wEGDVM without any conversion. wEGDVM is supplied to two convolutional layers followed by a recurrent network of three layers and an output layer, which finally gives the predicted class label. Fig. \ref{fig:DNN} illustrates the structure of our DNN classifier.

CNNs are non-fully-connected neural networks and are most commonly applied to analyzing visual imagery \cite{voulodimos2018deep}. They are efficient in capturing local relationships in matrix measures, such as boundaries and shapes in images \cite{zeiler2014visualizing}. The output of a CNN still has variable length, determined by the dimension of the input matrix measure. 

RNNs are a type of neural networks designed to learn from sequential input and are most widely used in natural language processing  \cite{lipton2015critical}. They are able to take inputs with variable length, and this capability is used in our model to process the output from CNNs. We use a special version of RNNs called bidirectional long short-term memory (bidirectional LSTM) networks \cite{hochreiter1997long, schuster1997bidirectional}. An LSTM unit is composed of a cell, an input gate, an output gate and a forget gate. The cell remembers values over arbitrary sequential intervals and the three gates regulate the flow of information getting into, extracting out of, and being remembered by the cell. This design equips the LSTM unit with the ability to preserve distant dependencies along a long sequence and thus capture the overall ``meaning/scheme'' of the sequence. The bidirectional version of LSTM further allows accessing information from both directions, which could bring additional power \cite{lipton2015critical}.

The CNNs-followed-by-LSTM-networks design can be quite suitable for protein structure classification. We hope to use CNNs to capture local protein structures such as turns in $\alpha$-helices and strands in $\beta$-sheets, and then let LSTM units move along the turns and strands to integrate the local protein structures into larger components such as $\alpha$-helices or $\beta$-sheets and finally give a global impression about the overall structure of the protein. The success of this approach relies on the ability of CNNs and LSTM networks to automatically learn what local structural patterns are useful and how to integrate them into higher level patterns that have direct implications for the class label. DNNs are known for their abilities to learn such patterns automatically during training \cite{lecun2015deep, rawat2017deep}.

Our model is implemented on TensorFlow 1.8 \cite{tensorflow2015-whitepaper}. The structure and other hyperparameters of the model are given in the Supplementary Materials.

\section*{Acknowledgments}
This work was supported by the National Institutes of Health [R01GM120733 to S.E., T.M., and J.L.].

\section*{Author Contributions}
H.G., T.M., and J.L. conceived the study; H.G., K.N., T.M., and J.L. proposed the methods for wEGDVM; H.G. and J.L. proposed the methods for DNN; H.G. and K.N. implemented the methods for wEGDVM; H.G. implemented the methods for DNN; H.G. analyzed the data, with help from K.N.; H.G. summarized the results, with help from T.M.; H.G. and J.L. drafted the manuscript; K.N., S.E., T.M., and J.L. revised the manuscript; T.M. and J.L. supervised the network science aspect of the study; J.L. supervised the machine learning aspect of the study.

\section*{Competing Interests statement}
The authors declare no competing interests.

\pagebreak
\bibliographystyle{unsrt}






\pagebreak

\section*{Figures}

\pagebreak

\pagebreak

\pagebreak

\end{document}